\documentclass{article}

\usepackage{microtype}
\usepackage{graphicx}
\usepackage{subfigure}
\usepackage{booktabs} % for professional tables
\usepackage{amsmath}
\usepackage{array}
\usepackage{multirow}
\usepackage{booktabs}  % 专业表格线
\usepackage{array}     % 增强表格功能
\usepackage{multirow}  % 合并单元格
\usepackage{adjustbox} % 智能缩放

\usepackage{longtable}
\usepackage{hyperref}

\usepackage[framemethod=TikZ]{mdframed} % 额外引入1
\usepackage[utf8]{inputenc} % allow utf-8 额外引入2
\usepackage{listings}

\usepackage{color}
\definecolor{keywordcolor}{rgb}{0.7, 0.1, 0.1}   % red
\definecolor{commentcolor}{rgb}{0.4, 0.4, 0.4}   % grey
\definecolor{symbolcolor}{rgb}{0.0, 0.1, 0.6}    % blue
\definecolor{sortcolor}{rgb}{0.1, 0.5, 0.1}      % green
\definecolor{errorcolor}{rgb}{1, 0, 0}           % bright red
\definecolor{stringcolor}{rgb}{0.5, 0.3, 0.2}    % brown

\usepackage[accepted]{icml2025}

\usepackage{amsmath}
\usepackage{amssymb}
\usepackage{mathtools}
\usepackage{amsthm}

\definecolor{darkpastelgreen}{rgb}{0.05, 0.65, 0.3}
\usepackage[capitalize,noabbrev]{cleveref}

\theoremstyle{plain}
\newtheorem{theorem}{Theorem}[section]

\newtheorem{corollary}[theorem]{Corollary}
\theoremstyle{definition}
\newtheorem{definition}[theorem]{Definition}

\theoremstyle{remark}

\usepackage[textsize=tiny]{todonotes}

\icmltitlerunning{KELPS: A Framework for Verified Multi-Language Autoformalization via Semantic-Syntactic Alignment}

\begin{document}
	
	\twocolumn[
	\icmltitle{KELPS: A Framework for Verified Multi-Language Autoformalization via Semantic-Syntactic Alignment}
	
	% It is OKAY to include author information, even for blind
	% submissions: the style file will automatically remove it for you
	% unless you've provided the [accepted] option to the icml2025
	% package.
	
	% List of affiliations: The first argument should be a (short)
	% identifier you will use later to specify author affiliations
	% Academic affiliations should list Department, University, City, Region, Country
	% Industry affiliations should list Company, City, Region, Country
	
	% You can specify symbols, otherwise they are numbered in order.
	% Ideally, you should not use this facility. Affiliations will be numbered
	% in order of appearance and this is the preferred way.
	\icmlsetsymbol{equal}{*}
	
	\begin{icmlauthorlist}
		\icmlauthor{Jiyao Zhang}{yyy,sch}
		\icmlauthor{Chengli Zhong}{yyy,sch}
		\icmlauthor{Hui Xu}{yyy}
		\icmlauthor{Qige Li}{yyy}
		\icmlauthor{Yi Zhou}{yyy,sch}
		%\icmlauthor{}{sch}
		%\icmlauthor{}{sch}
		%\icmlauthor{}{sch}
	\end{icmlauthorlist}
	
	\icmlaffiliation{yyy}{School of Information Science and Technology, University of Science and Technology of China, Hefei, China}
	\icmlaffiliation{sch}{USTC Knowledge Computing Lab}
	
	\icmlcorrespondingauthor{Jiyao Zhang}{ambitious\_777@mail.ustc.edu.cn}
        \icmlcorrespondingauthor{Yi Zhou}{yi\_zhou@ustc.edu.cn}
	
	% You may provide any keywords that you
	% find helpful for describing your paper; these are used to populate
	% the "keywords" metadata in the PDF but will not be shown in the document
	\icmlkeywords{Machine Learning, ICML}
	
	\vskip 0.3in
	]
	
	% this must go after the closing bracket ] following \twocolumn[ ...
	
	% This command actually creates the footnote in the first column
	% listing the affiliations and the copyright notice.
	% The command takes one argument, which is text to display at the start of the footnote.
	% The \icmlEqualContribution command is standard text for equal contribution.
	% Remove it (just {}) if you do not need this facility.
	
	%\printAffiliationsAndNotice{}  % leave blank if no need to mention equal contribution
	\printAffiliationsAndNotice{\icmlEqualContribution} % otherwise use the standard text.
	
	\begin{abstract}

    Modern large language models (LLMs) show promising progress in formalizing informal mathematics into machine-verifiable theorems. However, these methods still face bottlenecks due to the limited quantity and quality of multilingual parallel corpora.
    In this paper, we propose a novel neuro-symbolic framework KELPS (\textbf{K}nowledge-\textbf{E}quation based \textbf{L}ogical \textbf{P}rocessing \textbf{S}ystem) to address these problems. KELPS is an iterative framework for translating, synthesizing, and filtering informal data into multiple formal languages (Lean, Coq, and Isabelle).
    First, we translate natural language into Knowledge Equations (\textbf{KEs}), a novel language that we designed, theoretically grounded in assertional logic. 
    Next, we convert them to target languages through rigorously defined rules that preserve both syntactic structure and semantic meaning.
    This process yielded a parallel corpus of over 60,000 problems. Our framework achieves 88.9\% syntactic accuracy (pass@1) on MiniF2F, outperforming SOTA models such as Deepseek-V3 (81\%) and Herald (81.3\%) across multiple datasets. All datasets and codes are available in the supplementary materials.
    
	\end{abstract}

\begin{figure}[ht]
\begin{center}
\centerline{\includegraphics[width=\columnwidth]{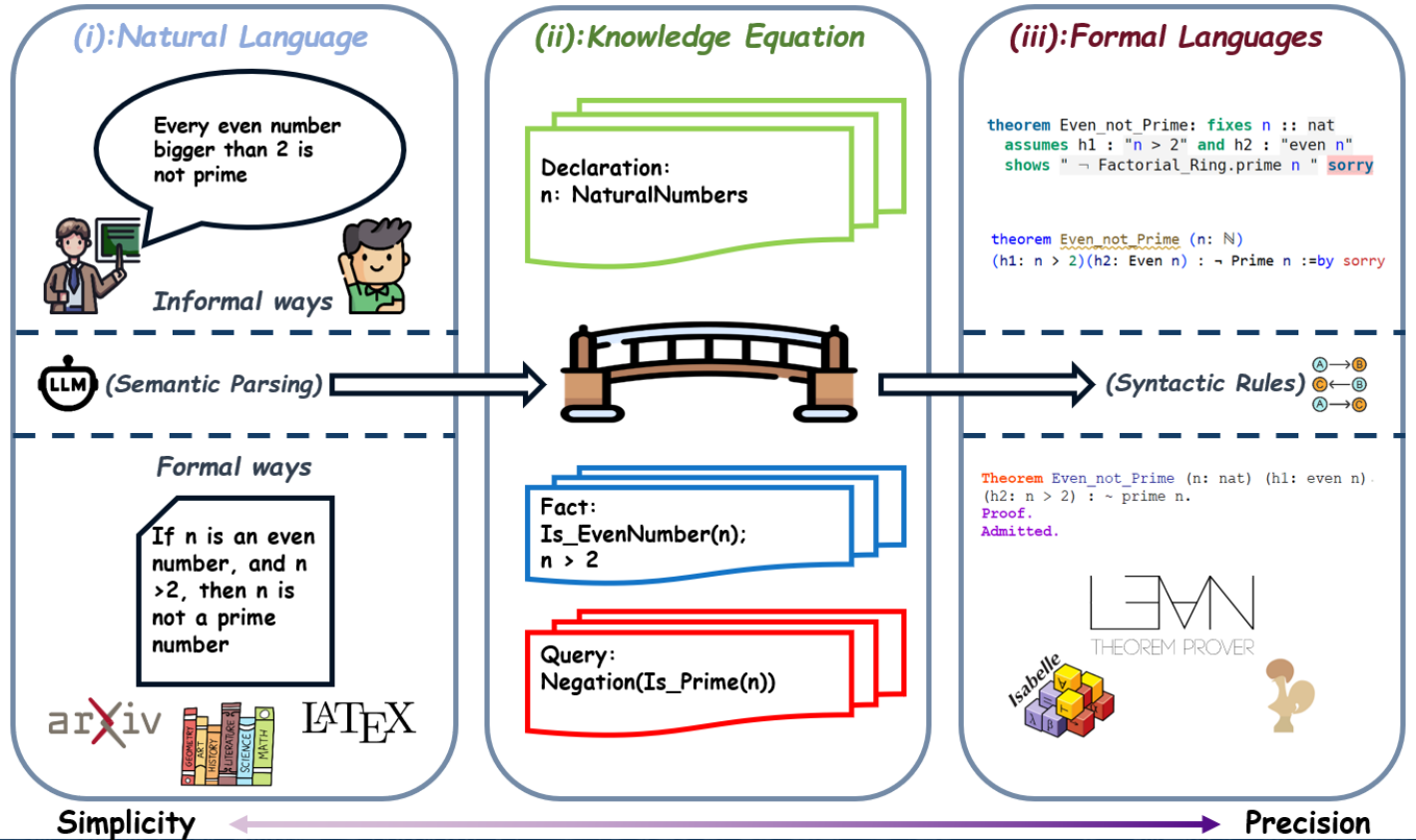}}
\caption{An overview of the relationships between natural language (NL), Knowledge Equations (KEs), and formal languages (FLs). KEs are extracted from NL through semantic parsing, then transformed into various FLs via specific syntactic rules.}
\label{alexample}
\end{center}
\vskip -0.35in
\end{figure}

	\section{Introduction}

    %The formalization of semantic meaning through rigorous logical systems has long been a longstanding objective of mathematicians from Leibniz and Hilbert to Wu Wen-Tsun \cite{wu2001mathematics}, since the inherent ambiguity of informal math hinder the verification of proofs and interdisciplinary collaboration, especially for advanced mathematics.

    Formalizing mathematical semantics as machine-verifiable codes has been a fundamental pursuit since Leibniz to Wu Wen-Tsun \cite{wu2001mathematics}, as informal statements' ambiguity impedes proof verification, particularly in advanced mathematics. Many sophisticated mathematical proofs span hundreds of pages and require extensive verification by experts, while such verification cannot eliminate minor errors or critical flaws.

    %The formalization of semantic meaning through rigorous logical systems has long been a longstanding objective of mathematicians from Leibniz and Hilbert to Wu Wen-Tsun \cite{wu2001mathematics}. As mentioned in Ganesalingam's book \cite{ganesalingam2013language}, the inherent ambiguity of informal mathematical discourse poses significant challenges—not only hindering interdisciplinary collaboration but also complicating the verification of proof correctness, especially for advanced mathematics.

    A rigorous formal system is therefore essential for unambiguous mathematical representation. Modern proof assistants have shown significant potential through landmark achievements such as the formalization of the Four-Color Theorem in Coq \cite{gonthier2008formal} and the Liquid Tensor Experiment \cite{scholze2022liquid} in Lean4. However, writing formal proofs is laborious and often challenging, highlighting the importance of autoformalization. This paper focuses specifically on statement autoformalization, a necessary prerequisite for proof autoformalization.

    %Modern proof assistants have demonstrated significant potential for this task, as evidenced by notable successes such as the formalization of the Four Color Theorem in Coq \cite{gonthier2008formal} and the Liquid Tensor Experiment \cite{scholze2022liquid} in Lean4. However, writing formal proofs is labor-intensive and often chanllaging, which highlights the importance of autoformalization. We only talk about autoformalizing statement in the next pages due to the huge gap between formal and informal reasoning.
    
    %Early attempts regarded this problem as machine translation problem \cite{wang2018first}. They trained neural models to translate LaTex-written text into the formal Mizar language. Recent studies including \cite{wu2022autoformalization}, \cite{patel2023new} and \cite{zhou2024don} focus on utilizing the in-context learning ability of LLMs, as LLMs have shown profound understanding of mathematical reasoning and multilingual translating. 

    Early attempts regarded this problem as a machine translation problem \cite{wang2018first}. They trained neural models to translate LaTex-written text into the formal Mizar language. Recent studies \cite{wu2022autoformalization} focus on utilizing the in-context learning ability of LLMs.
    
    %The most achievements such as \cite{gao2025herald} and \cite{liu2025atlas} tried fine-tuning models in a large-scale parallel NL-FL datasets. Despite promising results (\% 96.3 accuracy in MiniF2F) in MiniF2F, the scarity of large-scale general NL-FL pairs hampered bigger success.  Previous studies explored methods like translating NL problems into FL statements or back-translating FL corpus into NL statements to address this problem. 

    While fine-tuning models on large-scale parallel natural NL-FL datasets has achieved remarkable success, this approach remains constrained by the limited quantity and diversity of available NL-FL (natural language to formal language) pairs. Prior work has explored strategies such as translating natural language problems into formal language expressions \cite{ying2024lean} or back-translating formal language corpora into natural language statements \cite{gao2025herald}. Nevertheless, the quality of synthetically generated data, along with maintaining semantic alignment between FL and NL remains challenging.

    In this paper, We introduce KELPS (Knowledge Equation Linguistic Parsing System), a rule-based framework designed to translate natural language mathematical statements into multiple formal languages, including Lean4 \cite{moura2021lean}, Coq \cite{barras1999coq}, and Isabelle \cite{paulson1994isabelle}. As illustrated in Figure \ref{fig:custom_size}, KELPS consists of three core components. 
    
    \textbf{(1) Semantic Parsing}. The input natural language is first translated into an intermediate formal representation—the \textit{Knowledge Equation}. The formal definition and implementation details are in Section \ref{sec:3.1}.

    \textbf{(2) Syntactic Alignment}. The results in stage 1 are then parsed and converted into various formal languages via parallel complex rules. Compilers will help to check its correctness.

    \textbf{(3) Semantic Verification}. Formal statements generated in Stage 2, even though compiled successfully, still need semantic checks to be correct. We adopt the approach proposed in \cite{xin2025ape}, leveraging a LLM-as-a-Judge framework.

    We propose a data augmentation and expert iteration framework for progressive model enhancement. Our approach combines two key strategies: (1) A data synthesis module that efficiently generates various formalization data through randomized combinations of concepts and operators; (2) An expert iteration process where our model successfully parsed $>$60,000 formalized problems. The synthesized data set was used to train our KELPS translator model. Comprehensive evaluations demonstrate its superior performance, achieving \textbf{88.9\%} (pass@1) syntactic accuracy in the MiniF2F test, outperforming baseline models including Herald (81.3\%), DeepSeek-v3 (81\%) and Llama (61.4\%).

    Our main contributions are as follows.
    \begin{itemize}
        \item We propose a novel semantic parsing strategy that first transforms natural language into a simple intermediate representation. This representation enables flexible translation into multiple formal languages through pre-defined rules while preserving semantic fidelity.
        \item We propose a novel data synthesis strategy that enables precise atomic-level control over generated problems. We further provide an ontology that encompasses 40 core concepts and 180 operators.
        \item We introduce a dataset comprising 60,000 real-world and synthetic problems spanning K-12 and undergraduate-level mathematics. Our KELPS translator, trained on this dataset, shows substantial improvements over existing baselines on mainstream evaluation benchmarks (MiniF2F, FormalMATH).
    \end{itemize}

	\section{Related Work}
	
	\subsection{Formal System}
Natural languages exhibit inherent ambiguity in both textual and symbolic forms \cite{ganesalingam2013language}, complicating syntactic analysis and semantic extraction. This has motivated systematic efforts to formalize mathematical expressions through logical frameworks \cite{trybulec1989tarski, gordon2000lcf, carneiro2024lean4lean}. Existing approaches fall into two main distinct categories: controlled natural languages and formal language systems.

Controlled Natural Language (CNL) systems employ restricted natural language subsets with predefined grammars, as exemplified by Mizar \cite{grabowski2010mizar} and MathNat \cite{humayoun2010mathnat}. Recent advances include grammatical frameworks like GF \cite{ranta2004grammatical}, with GFLean \cite{pathak2024gflean} demonstrating direct text-to-Lean parsing.

%CNL Systems, such as Mizar \cite{grabowski2010mizar} and MathNat \cite{humayoun2010mathnat}, represent mathematical concepts through constrained natural language subsets with specified grammatical rules. Recently frameworks like Grammatical Framework \cite{ranta2004grammatical} gaining attention for structural analysis. For instance, GFLean \cite{pathak2024gflean} attempts to directly parse mathematical texts into Lean's formal language. 

Formal language systems (Lean, Coq) uniformly represent theorems through three core elements (\textbf{Declaration}, \textbf{Fact}, \textbf{Query}) despite differing in styles. To our knowledge, this work presents the first automated framework supporting multi-formal-language translation. Compared to existing formal languages, our knowledge equation framework achieves superior expressiveness and natural language alignment while maintaining simpler structure and better extensibility.

%Formal Language systems (Lean, Coq), while differing in their theoretical foundations and syntactic structures, universally allow theorem statements to be represented through the three components of knowledge equations (\textbf{Declaration}, \textbf{Fact}, \textbf{Query}). To our best knowledge, this constitutes the first automated framework capable of translating natural language into multiple formal languages. Compared to other approaches, our method offers superior expressiveness while maintaining closer alignment with natural language for enhanced interpretability.

%though it faces two critical limitations: (1) inefficiency due to natural language variability, and (2) insufficient expressiveness for higher-order logic. In contrast, our knowledge equation approach addresses these gaps by [key advantages: e.g., 'providing a unified intermediate representation that balances expressivity and automation']. (Example references could be added here for CNL systems.)
\begin{figure*}[t!]
  \centering
  \includegraphics[width=0.85\textwidth]{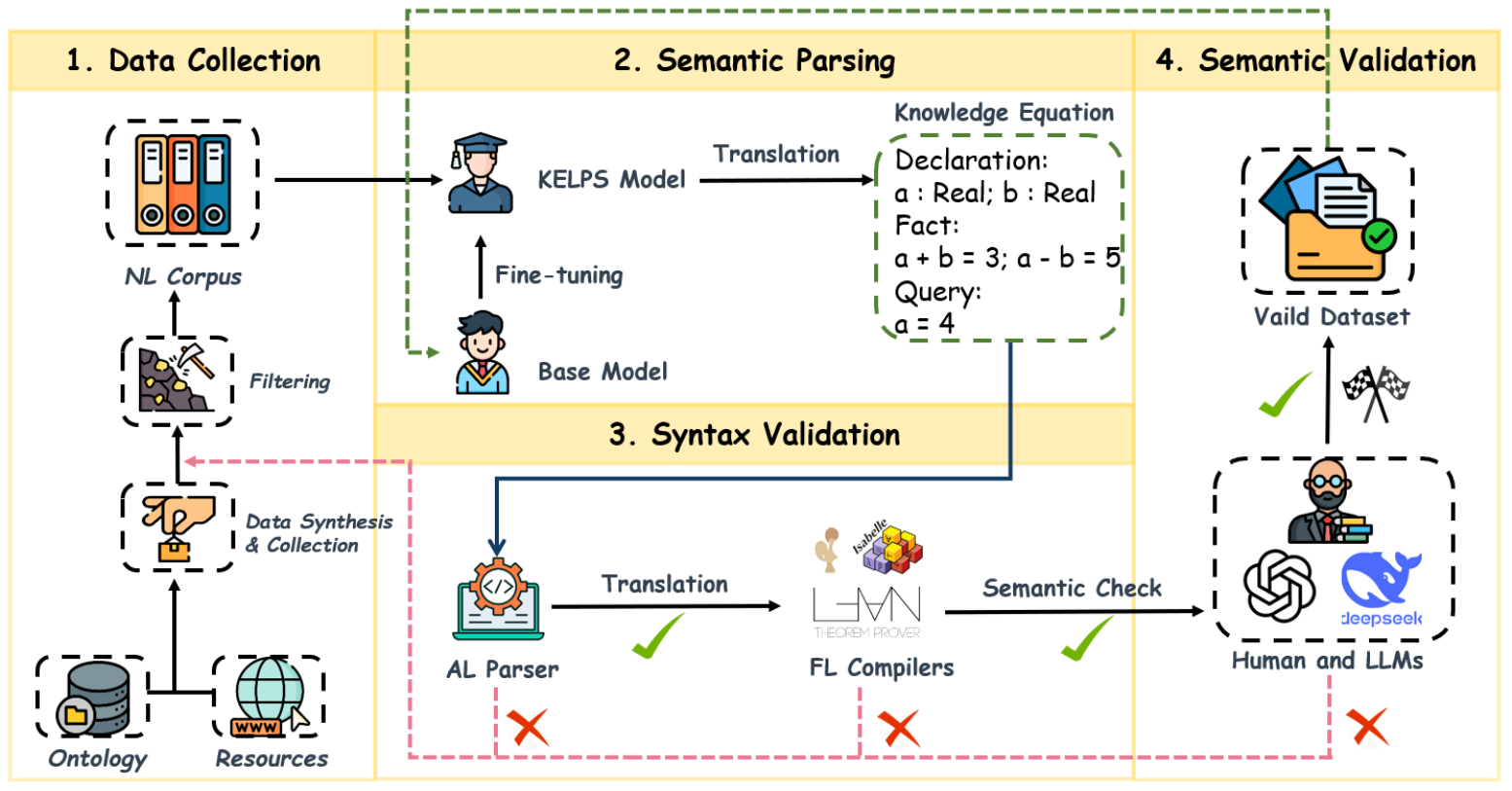}  % 95% 页面宽度，留白边
  \caption{Overview of Our Method. \textbf{(a) Data Collection}. We gather problems from various sources, including online resources and exercise sets, and construct an ontology library of relevant concepts and theorems. Through filtering and data synthesis strategies, we obtain a natural language (NL) corpus. \textbf{(b) Semantic Parsing}. We employ the KELPS model to perform semantic parsing, translating natural language problems into knowledge equations. The initial iteration of data is obtained through annotation. \textbf{(c) Syntax Validation}. The knowledge equations generated in \textbf{(b)} are validated by the AL Parser. Problems that pass validation are then converted into other formal languages via rule-based transformation. \textbf{(d) Semantic Validation}. Data that passes the compiler validation in the previous stage undergoes semantic review by both LLMs and human experts. Finally, the verified data is incorporated into the dataset, which is then used to continuously train the baseline model.}
  \label{fig:custom_size}
\end{figure*}
	\subsection{Autoformalization}

Autoformalization constitutes a specialized machine translation task that transforms natural language statements into formal representations while preserving semantic content and complying with target syntax requirements. Initial investigations explored neural approaches like \cite{wang2018first, cunningham2023towards}, demonstrating the feasibility of this paradigm.

Current research on LLM-based autoformalization primarily follows two dominant approaches: (1) few-shot in-context learning \cite{wu2022autoformalization, patel2023new, zhou2024don}, and (2) fine-tuning LLMs on NL-FL pairs \cite{lu2024formalalign, lu2024process, gao2025herald}. While the latter has shown promising results with \textbf{96}\% (pass@128) accuracy in MiniF2F \cite{zheng2021minif2f}, performance drops to just \textbf{16}\% (pass@128) on the College CoT benchmark, revealing the critical limitation of NL-FL data scarcity.

%However, the scarcity of large-scale NL-FL datasets remains a major bottleneck for further progress. As a result, only \textbf{16}\% (pass@128) accuracy in College CoT dataset.

A parallel research direction addresses the more challenging task of formal proof generation, where natural language proofs often diverge substantially from their formal counterparts. Current approaches include: (1) proof decomposition into draft skeletons with subsequent completion \cite{jiang2022draft, wang2023lego}, and (2) direct neural translation of informal proofs \cite{wang2024theoremllama, shao2024deepseekmath}. The first approach uses language models to complete proof steps within a structured framework, whereas the second aims for complete automated translation.

	\subsection{NL-FL Dataset Generation}
The scarcity of high-quality, large-scale natural language-formal language (NL-FL) parallel datasets remains a fundamental challenge in autoformalization research. Since hiring domain experts for annotation is expensive and inefficient, recent work has investigated leveraging large language models for scalable dataset generation. Existing approaches include two main categories:

The first line of work \cite{jiang2023multilingual, li2024numinamath, ying2024lean} exploits the wide availability of informal mathematical texts, using LLMs to translate NL statements into FL statements. Although this pipeline has yielded some large-scale datasets, it remains hampered by several limitations: the translation pipeline requires extensive post-processing, often produces low-quality statements, and faces challenges due to the scarcity of cutting-edge domain data.

%it is often cumbersome, requires multiple post-process steps, and suffers from low data quality and poor scalability.

The alternative paradigm \cite{wu2024lean, gao2025herald} initiates from formal language corpora, employing LLMs for backward translation to natural language.  While valuable, this approach remains fundamentally constrained by the scope and completeness of existing formal libraries.

Building upon existing work \cite{huang2024mustard, liu2025atlas} that generates diverse problems through randomized concept selection from a predefined concept library, we introduce a more controlled synthesis approach. Our method employs structured \textbf{Concept-Operator} templates to achieve three key advances: \textbf{(1)} ensuring comprehensive theorem coverage through a Concept-Operator library, \textbf{(2)} enabling easy-to-hard theorem synthesis via predefined templates, and \textbf{(3)} generating theorem diversity through various Concept-Operator combinations.

%To improve data diversity, previous approaches \cite{huang2024mustard, liu2025atlas} have attempted to generate diverse problems by randomly selecting concepts and difficulty levels from a predefined concept library, demonstrating notable effectiveness.  With this foundation, our method advances further by leveraging structured \textbf{Concept-Operator} templates to precisely control statement stylization while enabling flexible synthesis of theorems across a spectrum of difficulty levels from elementary to complex.

\section{Methodology}

In this section, we present our core methodologies for statement autoformalization and dataset construction. Section \ref{sec:3.1} establishes our theoretical foundations (\textbf{Assertional Logic} and \textbf{Knowledge Equations}). Section \ref{KELPS} presents the complete system architecture, and Section \ref{sec:3.3} details our synthetic data generation strategy.

%We propose KELPS, a unified framework for reliable translation from natural language to formal language, implemented as a three-stage pipeline (Figure 1):

%Semantic Parsing: Natural language inputs are mapped into an intermediate representation called Knowledge Equations (KEs), based on Assertional Logic.

%Rule-based Conversion: KEs are systematically transformed into formal code (e.g., Lean, Coq, Isabelle) via language-agnostic syntactic rules.

%Verification Optimization: LLM self-refinement and majority voting are used to ensure the correctness and robustness of the generated formalizations.

%Our framework introduces several key innovations:

%Semantic-Syntactic Decoupling: The use of KEs as an intermediate representation abstracts away formal language syntax, enabling cross-language semantic consistency (§3.1).

%Hybrid Data Generation: We integrate real-data parsing with synthetic augmentation to cover a broad spectrum of mathematical domains, from K-12 to graduate-level mathematics (§3.2).

%Extensible Parsers: KELPS provides a modular parser design that supports dynamic adaptation to both Assertional Logic and multiple target formal languages (§3.3).

%Experiments demonstrate that KELPS achieves state-of-the-art (SOTA) accuracy on MiniF2F and PutnamBench (§4). In addition, we construct the first large-scale multilingual formalization corpus ($>$ 30k NL–FL pairs), establishing a robust infrastructure for automated theorem proving and formal language translation.

\subsection{Assertional Logic and Knowledge Equation}
\label{sec:3.1}
We introduce Assertional Logic (AL), a knowledge representation system with formally specified syntax.

\subsubsection{An Introduction to Assertional Logic}

Assertional Logic (AL) \cite{zhou2017first}, is an extension of first-order logic with enhanced power as expressive as higher-order logic, while these representations are usually human-friendly.  

\begin{definition}
\label{def:inj}
The \textbf{Syntactic structure} of a given domain in AL is a tripe $\langle \mathcal{I}, \mathcal{C}, \mathcal{O} \rangle$. $\mathcal{I}$ is the collection of individuals, corresponding to objects in the domain. $\mathcal{C}$ is the collection of concepts, representing all sets of objects that have some properties in common. $\mathcal{O}$ is the collection of all operators, which acts among concepts and individuals like a function. 
\end{definition}

This structure has a natural correspondence with set theory, where individuals map to elements, concepts to sets, and operators to functions.

\begin{definition}
\label{def:assertion}
An \textbf{assertion} is the of form 
\begin{equation}
\begin{aligned}
a &= b 
\end{aligned}
\end{equation}
\end{definition}
where $a$ and $b$ are two terms. From a semantic perspective, it claims that the left and the right side refer to the same thing. A $term$
is an individual, either an atomic individual $a \in \mathcal{I}$ or the compound individuals $O(a_1,...,a_n)$. Where $O$ represents an operator on some individuals $ a_1, ..., a_n$.

%Despite analogous to first-order logic (FOL), there are instric %differences between AL and FOL, including  flexibility, expressivity %and extensibility. 

Building upon AL, we presented Knowledge Equations (KE) to represent all knowledge with the same form. There is an example in Fig \ref{alexample}. 

%and we will provide more details in Appendix \ref{BNF}. 

%There are three core components in AL: concepts, operators and individuals, with a corresponding relationship of 
%sets, functions and elements in set theory. In AL, all syntactic objects can be categorized into these definitions.
%Semantically, individuals, concepts and operators are interpreted as elements, sets and
%functions respectively in set theory
%Based on AL, we design a domain-specific formal language named Knowledge Equation, which serves as the foundation for expressing mathematical problems and reasoning tasks. A Knowledge Equation consists of three components:
%Declaration: defines the variables and associated concepts;
%Facts: encodes known assumptions or mathematical relationships;
%Query: specifies the question or goal to be inferred or proved.

%\begin{figure}[ht]
%\vskip 0.2in
%\begin{center}
%\centerline{\includegraphics[width=\columnwidth]{3}}
%\caption{Historical locations and number of accepted papers for International
%Machine Learning Conferences (ICML 1993 -- ICML 2008) and International
%Workshops on Machine Learning (ML 1988 -- ML 1992). At the time this figure was
%produced, the number of accepted papers for ICML 2008 was unknown and instead
%estimated.}
%\label{icml-historical}
%\end{center}
%\vskip -0.2in
%\end{figure}

\subsubsection{Translate KE into multiple language}

One of the biggest benefits of AL is that it could uniformly formalize every knowledge into assertions of the form $a= b$. Consequently, all mathematical assertions can be systematically translated into other formal languages, including but not limited to Lean and Coq. 

Unlike GFLean \cite{pathak2024gflean} which handles natural language  in its entirety, our method specifically targets assertions, concepts, and operators. The simplified syntax works with just a handful of core elements, enabling efficient translation to various formal languages.

In summary, a mathematical question can be divide into three parts: \textbf{Declartion}, \textbf{Fact} and \textbf{Query}. And we give each of their formal definitions.

\begin{definition}
\label{def:KEdeclaration}
\textbf{Declaration} part of a knowledge equation has the form of\begin{equation}
\begin{aligned}
\text{var : ConceptType}
\end{aligned}
\end{equation}
% $${var : ConceptType} $$
\end{definition}
where ``$\text{var}$" represents a free variable (belongs to Individual in AL) and ``$\text{ConceptType}$" is a defined concept. Its semantic meaning is similar to \text{``Assume $x$ is an integer ...''} and the syntactic part of ``fixes n :: int" in Isabelle or ``$(n : \mathbb{Z})$" in Lean4.

\begin{definition}
\label{def:KEfact}
\textbf{Fact} part of a knowledge equation has the form of \begin{equation}
\begin{aligned}
[\text{Assertion} 1; \text{Assertion}2; \text{Assertion}3...]
\end{aligned}
\end{equation}
\end{definition}
During the translation process, KELPS model will systematically capture all known information, (including inferable propositions such as ``Set $S$ is finite''), and formalize them as assertions. By leveraging the equivalence of a proposition $A \equiv (A = \text{True})$, this representation ensures seamless translation to various formal languages.

\begin{definition}
\label{def:KEquery}
\textbf{Query} part shares the syntactic structure with facts, but differs in their semantic meaning. \begin{equation}
\begin{aligned}
\text{Assertion}
\end{aligned}
\end{equation}

The assertions in the Query typically represent propositions requiring proof (for closed-form problems) or propositions we aim to investigate (for open-ended questions). For the later situation, we use \text{`` ? ''} as syntax sugar to replace the real item. This is just like \text{`` sorry ''} in Lean 4.
\end{definition}

We designed the Backus-Naur Form (BNF) of Knowledge Equations based on the formal definitions above. And by using ANTLR4, we implemented an extensible parser framework that automatically transforms Knowledge Equations into equivalent representations in target formal languages. The correctness is guaranteed by translation rules designed by human experts: \begin{equation}
\begin{aligned}
{C}_{\text{KE}} \mapsto {C}_{\text{TL}}, \quad {O}_{\text{KE}} \mapsto {O}_{\text{TL}}
\end{aligned}
\end{equation}
where ${C}_{\text{KE}}$ and ${O}_{\text{KE}}$ represents concepts and operators in KE, ${C}_{\text{TL}}$ and ${O}_{\text{TL}}$ represents the same semantic object in the target language.

%For terms beyond our boundary, we could fix it either by defining new terms in the target language or extending our ontology.

%Building on the theoretical foundation introduced above, we implement the BNF grammar and a corresponding parser for the Knowledge Equation language (see Figure XX). The parser translates a knowledge equation into an abstract syntax tree (AST). Due to the compact design and limited term types in Knowledge Equation, we can apply rule-based transformations to convert each component into a target formal language.

%For instance:
%In the Declaration block, we handle each declared concept using [specific transformation logic];
%In the Facts section, we process logical assertions through [another tailored mechanism];
%The Query is mapped to a formal goal statement, ready for theorem proving.
%For full technical details and transformation rules, please refer to the Appendix.

\subsection{KELPS Framework}
\label{KELPS}
%Our KELPS framework implements a rigorous three-stage verification pipeline to ensure both syntactic and semantic correctness during the formalization process. This section details the syntactic validation phase.
In this subsection, we present the KELPS framework (illustrated in Figure \ref{fig:custom_size}), which comprises three core components: Semantic Parsing, Syntactic Validation, and Semantic Validation. We will now describe each component in sequence.

%In this section, we elaborate on the procedure for syntax validation. The semantic parsing results obtained from Step 1 are passed into our custom parser for structural verification. The parsed output, once validated, is then forwarded to Lean for type-checking and formal verification.

\subsubsection{Semantic Parsing}
\label{sec 3.2.1}

Large-scale data annotation remains highly laborious. We develop an iterative pipeline that first fine-tunes DeepSeek-Math-7B-Base on \textbf{1,200} manually annotated examples (Fig. \ref{fig:custom_size}), then automatically processes unannotated data. The model's outputs undergo syntactic and semantic checks, with validated results expanding the training set through multiple refinement cycles. After seven iterations, our model processed $>$ 50K validated samples.

%Annotating large volumes of data is undoubtedly an extremely labor-intensive task. Here, we adopt a semi-automated pipeline to iteratively acquire training data. As shown in Figure \ref{fig:custom_size}, we first manually annotate \textbf{1,200} examples to fine-tune the DeepSeek-Math-7B-Base model to obtain a preliminary model capable of semantic parsing. Then, we apply this model to parse the remaining data and perform syntactic and semantic validation on the parsing results. The parts that pass the two verifications are added to the training set. This process is repeated to iteratively update the training set.

\subsubsection{Syntax Validation}
\label{sec 3.2.2}

%In this subsection, we detail the syntax validation pipeline. The semantic parsing results obtained in Step 1 are fed into our custom parser for structural and syntactic verification. Once the parsing succeeds, the validated output is passed to Lean for compilation and type-checking, ensuring formal correctness under the Lean proof assistant framework.

To ensure the syntactic correctness of the semantic parsing results obtained from Section \ref{sec 3.2.1}, we implement a two-stage validation process. 

We first perform grammar verification using our ANTLR4-based knowledge equation parser that guarantees strict compliance with formal specifications. Subsequently, we employ the target language's compiler to verify its final correctness.

For statements processed by the knowledge equation parser, approximately 80-90\% successfully pass validation through the target language compiler typically. The remaining cases primarily involve minor type conversion errors, which we analyze comprehensively in Appendix \ref{casestudy}.

\subsubsection{Semantic Validation}
\label{sec 3.2.3}

To ensure the semantic correctness of the semantic parsing results obtained from Section \ref{sec 3.2.2}, our evaluation framework incorporates insights from mainstream methods \cite{wu2022autoformalization,gao2025herald}. However, we observe that back-translation from formal to natural language fails to preserve semantic fidelity, thereby introducing extra measurement errors. Furthermore, the binary \textbf{True/False} classification criterion is insufficient for the precise measurement of semantic alignment in formalized expressions.

%First using a large language model to do back translation, translating the formalized results back into natural language to assess the preservation of meaning. Then we use the large language model to judge the semantic consistency between the back-translated statements and the original informal statements. 

We evaluate formalized statements through a graded alignment assessment framework \textbf{(0-5 scale)} with corresponding natural language expressions. Uncompilable statements are automatically assigned a score of 0. The evaluation is performed using DeepSeek-V3 (version 250324) with default parameters.

%To enhance robustness, we adopt a sample@3 strategy in the second stage: for each evaluation example, we collect three independent judgments from DeepSeek-v3. The final verdict is determined via majority voting, where semantic success is only confirmed if at least two 5-scores on semantic evaluation. All reported semantic evaluation results adhere to this protocol.

%Specifically, We employ the DeepSeek-v3 model (version 250324) for evaluation, using default parameter settings throughout. The specific prompts used are provided in Appendix C. 

To enhance robustness, we adopt a sample@3 strategy in the second stage: for each evaluation example, we collect three independent judgments from DeepSeek-v3. The final verdict is determined via majority voting, where semantic success is only confirmed if at least two 5 scores on semantic evaluation. All reported semantic evaluation results adhere to this protocol.

\subsection{AL-FL Data Generation}
\label{sec:3.3}
%This subsection details the strategy for constructing the KELPS dataset, a large-scale collection of NL-FL and AL-FL pairs comprising over 60,000 diverse samples supporting mainstream formal languages. 

This section presents the KELPS dataset construction methodology - a corpus of 60,000+ NL-FL and AL-FL pairs supporting major formal languages. 

Our pipeline first establishes a mathematics ontology, then employs dual generation strategies: (1) translation of natural language problems into AL representations, and (2) template-based synthesis of NL-AL pairs through a combination of templates. Implementation details follow in subsequent sections.

%We began by constructing a K12 mathematics ontology. Building upon this foundation, we adopted a dual-strategy approach to generate the dataset: translating NL problems into AL, and synthesizing NL-AL paired data through different templates. Detailed implementation methodologies are elaborated in the following subsections.

%We build a comprehensive ontology of K12-math, comprising total 40 concepts and 100 operators. Referring to this ontology, we construct a large-scale dataset covering main aspects of K12-math except geometry. We achieve this both by translating natural problems and syntheising new problems.

\subsubsection{ Building the Mathematical Ontology}

%We construct a comprehensive mathematical ontology tailored for the K-12 algebra domain. The ontology comprises XX Concepts and XX Operators, covering over 95 \% of typical problem types encountered in the K-12 mathematics curriculum.

We manually constructed a mathematics ontology covering most K12 and selected undergraduate-level mathematical topics, which comprises \textbf{6} major topics, \textbf{40} core concepts, and \textbf{180} operators. This ontology development effort required approximately three weeks of work by two mathematical graduate students.

The complete specifications and structural details of the ontology are provided in Appendix \ref{ontology}.

\subsubsection{Collection and Processing Data}

This module discusses two main steps in building our dataset from NL problems ---- problem collection, filtering, and translation.

\textbf{NL Problems Collection}

The Numina \cite{li2024numinamath} dataset represents the largest and most comprehensive open-source collection of K12 mathematics materials, incorporating diverse sources ranging from AIME competition problems to Chinese K12 curriculum content. In this study, we selected Numina as our primary reference dataset due to its unique coverage that subsumes content from numerous other mathematical datasets.

%For natural language data, we primarily leverage the Numina dataset, currently the largest known collection of K-12 mathematical word problems, containing approximately 380,000 questions. Through rigorous filtering and formalization, we curate a high-quality subset of 50,000 NL-FL aligned examples.

\textbf{Filtering and Translation}

We first filtered out unsupported problems from the Numina dataset, including out-of-ontology mathematical questions, retaining approximately 100,000 items. Subsequently, we removed problems unsuitable for formalization, such as questions with graphical representations. Finally, we yielded a corpus of 70,000 problems for translation.

We adopted an expert-iterative approach for continuous problem parsing and successfully translated \textbf{50,000} problems after seven iterations. 

%More details are presented in Section 4.3.

\begin{table*}[t]
\caption{Evaluating Performance Across Different Models and Datasets. We highlight the top-performing results for \textcolor{red}{syntactic accuracy} in red and those for \textcolor{darkpastelgreen}{semantic accuracy} in green.       Syntactic accuracy is determined by whether the code passes compiler verification, while semantic accuracy scores are obtained through LLM majority voting. The version of Deepseek-V3 used is \textbf{DeepSeek-V3-0324}. Given the substantial volume of problems, we selectively sampled \textbf{200} questions from FormalMATH and \textbf{300} from Numina-Hard for experiments. }
\label{table1}
\vskip 0.15in
\begin{center}
\begin{small}
\begin{tabular}{ccccccccc}
\toprule
\multirow{2}{*}{\textbf{Model}} & \multicolumn{2}{c}{\textbf{MiniF2F}} & \multicolumn{2}{c}{\textbf{MiniF2F-Isabelle}}  & \multicolumn{2}{c}{\textbf{Numina-Hard}} & \multicolumn{2}{c}{\textbf{FormalMATH}} \\ \cmidrule{2-9} 
 & Syntax & Semantic & Syntax & Semantic & Syntax & Semantic & Syntax & Semantic \\ 
\midrule
DeepSeek-V3 (671B) & 81\% & \textcolor{darkpastelgreen}{4.00} & \textcolor{red}{56.5}\% & \textcolor{darkpastelgreen}{2.74} & 82\% & \textcolor {darkpastelgreen}{4.02} & 58.3\% & 2.90 \\
Herald (7B) & \textcolor{red}{81.3\%} & 3.58 & -- & -- & \textcolor{red} {85.5\%} & 3.30 & \textcolor{red}{73.8\%} & \textcolor{darkpastelgreen}{3.11} \\
LlaMa-3 (8B) & 61.4\% & 2.63 & 54.3\% & 2.27 & 62.3\% & 2.58 & 37.7\% & 1.56 \\
\textbf{KELPS} (7B) & \textcolor{red}{\textbf{88.9\%}} & \textcolor{darkpastelgreen}{\textbf{4.05}} & \textcolor{red}{\textbf{82.2\%}} & \textcolor{darkpastelgreen}{\textbf{3.56}} & \textcolor{red}{\textbf{94.3\%}} & \textcolor{darkpastelgreen}{\textbf{4.49}} & \textcolor{red}{\textbf{74.3\%}} & \textcolor{darkpastelgreen}{\textbf{3.29}} \\
\bottomrule
\end{tabular}
\end{small}
\end{center}
\vskip -0.1in
\end{table*}

\subsubsection{Synthesis Strategies}

%Distinct from previous methods, we introduce a novel synthesis strategy for generating formal language data. We design compositional templates by combining various Concepts, Operators, and symbolic tokens to serve as generation patterns. These templates range from elementary tasks—such as computing the greatest common divisor of two numbers—to complex competition-style problems that require intricate reasoning under specified constraints. We manually construct a set of XX templates, and employ DeepSeek-v3 to automatically generate XX synthesized examples based on these templates.

This subsection presents our synthetic data generation framework, which consists of two key components: template creation based on our mathematical ontology, and a systematic template combination strategy.

\textbf{Templates Creation}

Using the strong in-context learning capabilities of LLM, we found that providing just 3-shot examples was sufficient for the model to correctly learn patterns of target concepts and operators. For our template-based generation tasks, DeepSeek-v3 \cite{liu2024deepseek} achieved a 90\% syntactic accuracy. 

In addition to templates derived from individual concepts, we extract specialized templates from concrete problem instances. For example, the problem ``Find all integers such that [a is a prime and a $<$ 15]'' is abstracted into a template where the condition ``[a is a prime and a $<$ 15]'' is replaced with a generic ``[property]'' placeholder. This approach enhances the model’s capacity to generate diverse problem variations.

%%%
%\begin{figure}[ht]
%\vskip 0.2in
%\begin{center}
%\centerline{\includegraphics[width=\columnwidth]{4}}
%\caption{Historical locations and number of accepted papers for International
%Machine Learning Conferences (ICML 1993 -- ICML 2008) and International
%Workshops on Machine Learning (ML 1988 -- ML 1992). At the time this figure was
%produced, the number of accepted papers for ICML 2008 was unknown and instead
%estimated.}
%\label{icml-historical}
%\end{center}
%\vskip -0.2in
%\end{figure}
%%%

\textbf{Templates Combination}

However, this approach tended to produce problems with limited diversity, and the model occasionally made errors due to incomplete conceptual understanding. To address these limitations, we developed a composite template strategy that systematically combines templates of varying complexity - from basic concept applications to advanced problem types. We ultimately constructed a corpus of 50+ high-quality templates, significantly improving both the diversity and complexity of generated problems.

The complete set of prompting templates employed to generate synthetic data is provided in Appendix \ref{prompt}.

%The complete pipeline is illustrated in Figure 3, with complementary case studies in Appendix C.

%A detailed discussion about the effect of synthetic data appears in Section 4.3

\section{Experiments}
\label{sec: 4}

We conduct a comprehensive series of experiments to evaluate both the performance of the KELPS translator and the quality of the KELPS dataset. Section \ref{sec: 4.1} details the experimental setup and configurations. Section \ref{sec: 4.2} presents the main results of multiple benchmarks. Section \ref{sec: 4.3} provides an ablation study to analyze the effect of different components. Finally, Section \ref{sec: 4.4} offers a further analysis of the influencing factors behind the results.
	
\subsection{Experimental Setup}
    \label{sec: 4.1}
\textbf{Fine-tuning}.  We employ DeepSeek-Math-7B-Base as our base model and conduct supervised fine-tuning on the KELPS dataset using a full-parameter training approach. 

\textbf{Dataset}. To evaluate the performance of our system, we conduct comparative benchmarks across three mathematical datasets: MiniF2F, FormalMATH, and Numina-Hard. These established benchmarks comprehensively cover olympiad/undergraduate mathematics through diverse problem types, ideal for formalization testing.

\begin{itemize}
\item \textbf{MiniF2F}. \cite{zheng2021minif2f} A widely adopted multilingual benchmark for auto-formalization tasks, comprising K12-level mathematical problems. In this work, we only evaluate its test set.
\item \textbf{FormalMATH}. \cite{yu2025formalmath} A formalized mathematical benchmark that spans Olympic competitions and undergraduate-level problems in multiple domains. We randomly select 200 problems across various mathematical domains for evaluation.

\item \textbf{Numina-Hard}. The Numina-Hard dataset comprises 300 challenging problems randomly selected from the KELPS dataset.
\end{itemize}

Due to the comprehensiveness and practicality of \textbf{Mathlib} \cite{blokpoel2024mathlib}, we adopt Lean4 as our primary formalization language. All experiments are conducted in Lean4 by default. In addition, we evaluate the \textbf{NL} to \textbf{Isabelle} translation on the MiniF2F benchmark. Herald's results \cite{gao2025herald} were excluded as their experimental setup diverged from our task requirements.

% as shown in Appendix xx.

%with the following configuration: The model was trained with a batch size of 16 samples per device (using 4 gradient accumulation steps), using the AdamW optimizer with bfloat16 precision. We adopted an initial learning rate of 4.0e-5 with cosine scheduling, trained for 15 epochs, and applied a warmup ratio of 10\% of the total training steps.

%\textbf{Baseline}. To evaluate the performance of our system, we benchmark different models on two benchmark datasets: MiniF2F and KELPS-Dataset. 

%All experiments are conducted under the same decoding strategies to ensure fair comparison: \textbf{Top-p} is set to \textbf{0.9}, \textbf{Temperature} is set to \textbf{0.7} to improve output diversity, and \textbf{Max tokens} is set to \textbf{256}. Unless otherwise stated, all models are prompted using the same KELPS translation pipeline.

\textbf{Experimental Process}. To validate our model's capabilities, we employ the experimental pipeline illustrated in Fig \ref{fig:custom_size}. It comprises three core steps: \textbf{semantic parsing}, \textbf{syntax validation}, and \textbf{semantic validation}, with detailed descriptions provided in Section \ref{sec:3.3}. For the Herald model, we rigorously maintain the original configuration reported in \cite{gao2025herald}. The implementation details and hyperparameter settings for all other models are provided in Appendix \ref{appendixB}.

All KELPS experiments used NL-Lean4 fine-tuning and translated directly into Lean4 in this chapter, except for \textbf{MiniF2F-Isabelle}, which employed NL-AL training followed by rule-based Isabelle translation.

Our experimental environment utilizes Lean v4.19.0 (with Mathlib4 of the same version), and Isabelle 2025 (March 2025). The header files and executable code used in our experiments are included in our supplementary materials.

    \begin{table*}[t]
\caption{We compare the performance of KELPS trained on datasets of varying scales and composition ratios across different benchmark datasets, where the 'Dataset ratio' denotes the proportion of \textbf{informal data} from natural language problems to model-generated \textbf{synthetic data} in the training corpus.}
\label{table2}
\vskip 0.15in
\begin{center}
\begin{small}
\begin{tabular}{cccccccc}
\toprule
\multirow{2}{*}{\textbf{Dataset Ratio}} & \multirow{2}{*}{\textbf{Dataset Size}} & \multicolumn{2}{c}{\textbf{MiniF2F}} & \multicolumn{2}{c}{\textbf{Numina-Hard}} & \multicolumn{2}{c}{\textbf{FormalMATH}} \\ \cmidrule(lr){3-4} \cmidrule(lr){5-6} \cmidrule(lr){7-8} 
 & & \textbf{Syntax} & \textbf{Semantic} & \textbf{Syntax} & \textbf{Semantic} & \textbf{Syntax} & \textbf{Semantic} \\ 
\midrule
1 : 0 & 14k & 81.6\% & 3.75 & 91.7\% & 4.33 & 60.1\% & 2.54 \\
1 : 0.5 & 21k & 87.6\% & 4.08 & \textcolor{red}{\textbf{94.5\%}} & \textcolor{darkpastelgreen}{\textbf{4.51}} & 70.3\% & 3.14 \\
1 : 1 & 28k & 88.4\% & \textcolor{darkpastelgreen}{\textbf{4.08}} & 93.4\% & 4.49 & 70.9\% & 3.25 \\
1 : 1.5 & 35k & \textcolor{red}{\textbf{88.9\%}} & 4.05 & 94.3\% & 4.49 & \textcolor{red}{\textbf{74.3\%}} & \textcolor{darkpastelgreen}{\textbf{3.29}} \\
\bottomrule
\end{tabular}
\end{small}
\end{center}
\vskip -0.1in
\end{table*}

\subsection{Main Results}
\label{sec: 4.2}

In this section, we present a systematic comparison between the KELPS model and other baseline models across various benchmark datasets. Our evaluation framework assesses two critical dimensions of model performance: (1) \textbf{syntactic accuracy}, measuring formal correctness, and (2) \textbf{semantic accuracy}, evaluating meaningful correspondence to mathematical truth.

\textbf{Syntactic Accuracy} quantifies the formal correctness of model outputs by measuring their ability to pass automated compiler verification. The overall pass rate is equal to the ratio of \textbf{compiler-valid statements} to \textbf{all problems}.

\textbf{Semantic Accuracy} assesses the mathematical equivalence between formally verified statements and their original natural language formulations, following the evaluation protocol established in \ref{sec 3.2.3}. The semantic scores reported in Table \ref{table1} represent the \textbf{average performance} across all problems. Problems that fail to compile are assigned a score of \textbf{zero}.

%To evaluate the performance of our model, we conducted comprehensive tests comparing KELPS with several models in similar settings. We evaluated our accuracy using pass@k \cite{chen2021evaluating} evaluation with k = 1 and 8. A sample is considered correct only if it passes both semantic verification and syntax checking. Among the k generated samples, if at least one passes these checks, it is counted as valid. 

%Due to limited time and computational resources, and considering that both the Herald and KELPS models already demonstrated satisfactory performance at k=8, we only report results for k=1 and 8. 

%For our experimental evaluation, we employed two benchmark datasets: miniF2F \cite{zheng2021minif2f} and KELPS-Test. In this work, we specifically utilized only the test set of miniF2F for our experiments.

%The KELPS-Test dataset comprises 900 problems randomly sampled from our full KELPS collection, designed to assess model performance on K-12 level mathematical problems. To ensure comprehensive evaluation, we stratified these problems into three difficulty levels (\textbf{easy}, \textbf{medium}, and \textbf{hard}), with 300 problems in each category. Furthermore, we implemented careful topic control during sampling to guarantee adequate coverage across all math problems.

As summarized in Table \ref{table1}, our experimental results demonstrate that the KELPS translator maintains robust performance across all evaluation metrics and datasets. Notably, while other models exhibit significant accuracy gaps between Lean and Isabelle formalizations, the KELPS translator's consistent performance confirms its capability for cross-formal-language translation. 

We also note the outstanding performance of the Herald model on the FormalMATH dataset, which may be attributed to its training on extensive advanced Lean4 statements. We will delve deeper into this phenomenon in Section \ref{sec: 4.4}.

The results demonstrate that the KELPS translator excels at formalizing natural language sentences into various formal representations. A case study of the formalization results of the KELPS translator is shown in Appendix \ref{casestudy}.

	\subsection{Ablation Study}
    \label{sec: 4.3}
In this subsection, we conduct ablation studies on the synthetic data approach, primarily investigating the impact of synthetic data quantity and quality on model performance. We compare the results under different configurations. All models were trained using a subset of the KELPS dataset containing approximately \textbf{15,000} Numina-Basic NL-AL pairs parsed from Numina combined with various categories of synthetic data. To systematically evaluate the effects, we controlled two key factors: (1) dataset diversity and quality by varying the templates used for synthetic data generation, (2) the mixing ratio between synthetic and authentic data samples.

As shown in Table \ref{table2}, the experimental results demonstrate that our synthetic data augmentation approach significantly improves model performance. We observe a nonlinear relationship between training data scale and model performance. When the dataset is small ($<$ 20K samples), increasing its size significantly improves both semantic understanding and grammatical accuracy. However, beyond a certain threshold, further scaling yields diminishing returns. This implies that prioritizing \textbf{high-quality, diverse training data} may be more effective than simply pursuing larger quantities.

%More importantly, we observed that \textbf{data quality plays a substantially more critical role than mere quantity} during the experiments. Specifically, with only 15,000 high-quality Numina-derived samples, the model already achieved an accuracy of XX on the MiniF2F benchmark. In contrast, incorporating partially verified synthetic data samples led to noticeable performance degradation.

Due to the limited variety of operators and homogeneous combinations in the training data, the model often fails to handle infrequent or structurally complex problems. To address this limitation, we employed three undergraduate students to systematically analyze mathematical templates from competition problems and online K12 mathematical resources. These templates focus on specific forms and categories of problems, significantly alleviating the scarcity of relevant data.

Compared to the limited data variety in Numina, synthetic data significantly enriches training diversity—yielding notable performance gains by incorporating only 7,000 additional problems. However, constrained by template rigidity and limited conceptual combinations, further scaling fails to produce substantial breakthroughs. This limitation is particularly evident on competition-level datasets (e.g., FormalMATH), where synthetic data complexity remains inadequate, resulting in lower performance relative to simpler benchmarks. Therefore, future research should prioritize enhancing data quality over mere quantity. To achieve this objective, establishing an effective evaluation metric for problem complexity is both essential and urgent.

%Based on these templates, we synthesized analogous problems. As evidenced by Table \ref{table2}, incorporating these more sophisticated templates improved the model's accuracy to XX. This finding further substantiates the critical impact of data quality on model performance. 

%This subsection analyzes the individual contribution of the KELPS framework through a series of controlled ablation experiments. 
%We investigate the effect of: Training with only real-world data \& Training with only synthetic (template-generated) data \& Training with KELPS-generated data.

%The results are shown in Table 2. From the table, we observe that the model trained solely on synthetic data yields the poorest performance on the MiniF2F and KELPS datasets, with a decrease of xx percentage points compared to the model trained exclusively on real data. This performance drop may be attributed to the limited diversity of the generated data templates. In contrast, the model trained on a combination of real and synthetic data achieves the best performance, 
%which demonstrates that incorporating both real and synthetic data, guided by the KELPS formalism, significantly enhances model generalization and logical consistency.

\subsection{Analysis}
\label{sec: 4.4}
In this subsection, we provide an in-depth analysis of the experimental outcomes and identify key factors that influence model performance. These factors represent the primary targets for improvement in our subsequent research.

\textbf{Effectiveness of KE Translation}: We will discuss the impact of introducing Knowledge Equations. Specifically, the task of autoformalization can be regarded as an alignment between natural language and formal language.
\begin{corollary} This alignment can be further categorized along the following two dimensions:
$$\text{AG}(\text{NL}, \text{FL}) = \text{AG}_{\text{syn}}(\text{NL}, \text{FL}) + \text{AG}_{\text{sem}}(\text{NL}, \text{FL})$$
\end{corollary}
where $\text{AG}$ denotes the overall alignment gap between the informal statement (\text{NL}) and the formal statement (\text{FL}). 
$\text{AG}_{\text{syn}}$ represents the \textit{syntactic alignment gap},
while $\text{AG}_{\text{sem}}$ denotes the \textit{semantic alignment gap}.
\begin{corollary} The term $\text{AG}_{\text{syn}}$ primarily depends on the syntactic rules and stylistic conventions of the FL. And $\text{AG}_{\text{syn}}$ coule be decomposed into two subcomponents
$$\text{AG}_{\text{syn}}(\text{NL}, \text{FL}) = \text{AG}_{\text{syn}}(\text{NL}, \text{AL}) + \text{AG}_{\text{syn}}(\text{AL}, \text{FL})$$
\end{corollary}
where the $\text{AG}_{\text{syn}}$ part can be automatically resolved by our rule-based parser

Table \ref{table1} shows $\text{AG}(\text{NL}, \text{FL})$ varies significantly across formal languages, causing DeepSeek-v3's performance to drop from 81\% to 56.5\% on MiniF2F. In contrast, our method only requires $\text{AG}(\text{NL}, \text{AL})$ mapping, thus with relatively consistent results (88.9\% and 82.2\%)

\textbf{Quality of Alignment between NL-FL pairs}:  
Since the semantic verification module that relies on a large language model as a referee is not completely reliable, instances of misalignment between NL and FL persist in the training dataset. Common misalignment patterns include misinterpretation of problem semantics and omission of problem-specific constraints. 

Our preliminary experiments revealed that semantically erroneous data can mislead models to generate incorrect responses. To address this issue, We employed a rigorous manual verification process to curate a high-quality subset of 15,000 precisely aligned NL-FL pairs for experiments.

%Therefore, we employed a rigorous manual verification process to curate a high-quality subset of 15,000 precisely aligned NL-FL pairs as the final training dataset. As shown in Appendix \ref{prompt}, when we remove this part from the training set, the final performance of both syntax and semantic is greatly improved. This result further implies that \textbf{misaligned samples} may degrade the model's performance.

\textbf{The Diversity and Coverage of the KELPS Ontology}:  
%Since the KELPS ontology does not include the geometry domain, we cannot correctly parse the relevant questions in the miniF2F dataset. If we want to expand the formalization to more fields, the first problem to be solved is the expansion of the ontology library.
Due to limited time and resources, we only modeled core mathematical concepts in natural language. While these atomic concepts are theoretically sufficient to represent most mathematical knowledge, our experiments revealed that the model tends to \textbf{self-construct undefined operators} (shown in Appendix \ref{casestudy}). These operators failed syntax validation because of the absence of corresponding transformation rules.

Therefore, to enhance both the practical utility and expressive power of Knowledge Equations, our future research will focus on comprehensive semantic modeling of natural language in its entirety.

%Due to the unique grammatical rules of AL, specific numbers (such as $\sqrt{2}$, $1$, $4.3$) appearing in the text will not be declared to belong to which type, while formal languages such as Lean, Coq, and Isabelle require the type of each variable to be clear, which brings a challenge how to map each specific number to the correct type. This is also the key reason why we lose accuracy when transforming from knowledge equation to other formal languages.

%%%%%%%%%%%%%%%%%%%%%%%%%%%%%%%%%%%%%%%%%%%%%%%%%%%%%%%%%%%%%%%%%%%%%%%%%%%%%%%
	%\section{Discussion}
	%This Part needs to talk about the big picture.
	%\subsection{Limitations}  

	%\subsection{Future Work}  
%%%%%%%%%%%%%%%%%%%%%%%%%%%%%%%%%%%%%%%%%%%%%%%%%%%%%%%%%%%%%%%%%%%%%%%%%%%%%%%

\section{Conclusion}
In this work, we propose KELPS, a novel three-stage framework for autoformalization. Our method introduces an intermediate representation—Knowledge Equation, which translates natural language into multiple formal languages. This representation aligns more closely with natural language, thus improving model accuracy, while the expert-crafted transformation rules guarantee syntactic correctness.

%Compared to direct NL-FL translation, Knowledge Equations are semantically rich yet structurally closer to natural language, making them easier to generate while remaining flexible enough to capture complex mathematical semantics. This intermediate representation can be easily adapted to various domains and supports multiple downstream formal languages, including Lean, Coq, and Isabelle.

We introduce a large-scale parallel dataset comprising over 60,000 NL-FL pairs spanning distinct mathematical subfields, with multi-language support (Lean, Coq, Isabelle). Additionally, we propose an LLM-based synthetic data generation strategy that controls difficulty levels and targets specific concepts/operators, effectively enhancing data diversity. Fine-tuning the KELPS translator on this dataset achieved 88.9\% syntactic accuracy on the MiniF2F dataset, outperforming SOTA models like Deepseek-V3 81\% and Herald 81.3\% on MiniF2F.

In summary, this work makes three contributions: the design of Knowledge Equations as a novel formal language, including its complete BNF specification and parser implementation; the release of KELPS Dataset, a large-scale multilingual formal language dataset; and the development of a high-performance model that achieves new SOTA accuracy. Due to their simplified formalism, Knowledge Equations demonstrate strong potential as both an educational and research-oriented formal language, despite limitations in automated type conversion between multiple concepts.

In the future, we plan to extend and refine this methodology along two key dimensions: expanding Knowledge Equations' coverage to advanced mathematical domains, and strengthening its theoretical foundations to address type system conversion errors. We posit that Knowledge Equations possess the potential to emerge as a universal, machine-learnable language.

\section{Impact Statement}

This paper presents work whose goal is to advance the field
of Machine Learning. There are many potential societal
consequences of our work, none which we feel must be
specifically highlighted here.

	\bibliographystyle{icml2025}
	\bibliography{example_paper}
	
	%%%%%%%%%%%%%%%%%%%%%%%%%%%%%%%%%%%%%%%%%%%%%%%%%%%%%%%%%%%%%%%%%%%%%%%%%%%%%%%
	%%%%%%%%%%%%%%%%%%%%%%%%%%%%%%%%%%%%%%%%%%%%%%%%%%%%%%%%%%%%%%%%%%%%%%%%%%%%%%%
	%APPENDIX
	%%%%%%%%%%%%%%%%%%%%%%%%%%%%%%%%%%%%%%%%%%%%%%%%%%%%%%%%%%%%%%%%%%%%%%%%%%%%%%%
	%%%%%%%%%%%%%%%%%%%%%%%%%%%%%%%%%%%%%%%%%%%%%%%%%%%%%%%%%%%%%%%%%%%%%%%%%%%%%%%
	\newpage
	\appendix
	\onecolumn
\begin{table}[h]
\caption{\fontsize{9pt}{10pt}\selectfont Representative Ontology Examples}
\label{tab:ontology}
\vskip 0.15in
\begin{center}
\begin{small}
%\begin{sc} % ⛔ 删除 small caps
\begin{tabular}{llp{5.55cm}}
\toprule
\textbf{Domain} & \textbf{Concept} & \textbf{Operator} \\
\midrule
\multirow{7}{*}{Numbers} 
& \multirow{2}{*}{\centering Real} & Abs, Sqrt, Log, NaturalLog, Sqrt, Cos, Sin, Tan, Get\_Number\_Round, Exp, Is\_Real \\
 \cmidrule(lr){2-3}
& \multirow{2}{*}{\centering Integers} & Get\_GCD, Is\_OddNumber, Get\_Remainder, Get\_InversedMod, Get\_LCM,  \\
 \cmidrule(lr){2-3}
& \multirow{3}{*}{\centering NaturalNumbers} & Factorial, Get\_Combination, Is\_Coprime, Is\_Prime, Get\_Digit\_Number, Get\_DigitSum, Get\_DigitProduct, Get\_DigitCount  \\
\midrule
\multirow{2}{*}{Polynomial} 
& \multirow{2}{*}{\centering Polynomial} & Get\_PolyTerm,  Is\_PolyFactor, Get\_Polyroots, Get\_PolyDegree, Get\_Term\_Coefficient, \\

\midrule
\multirow{3}{*}{Function} 
& \multirow{3}{*}{\centering Function} & Get\_Function\_Range, Get\_Function\_Zeroes, Get\_Inverse\_Function, Get\_Function\_Value, Get\_Function\_Minimum, Is\_Bijection \\

\midrule

\multirow{3}{*}{Set} 
& \multirow{3}{*}{\centering Set} & Set\_Cardinality, Set\_Union, Set\_Difference, Set\_Intersection, Build\_Set, Get\_Set\_Sum, Get\_Set\_Maximum, Get\_Set\_Minimum \\

\midrule

 \multirow{3}{*}{Sequence} 
& \multirow{3}{*}{\centering Sequence} & Is\_GeometricSequence, Get\_CommonRatio, Is\_ArithmeticSequence, Get\_Sequence\_Sum, Get\_CommonDifference \\

\midrule

 \multirow{2}{*}{Special} 
& \multirow{2}{*}{\centering --------} & ForAll, Exists, Get\_Prod, Solve\_equation, Get\_Sum , Negation, Range \\

\bottomrule
\end{tabular}
%\end{sc} % ⛔ 删除 small caps
\end{small}
\end{center}
\vskip -0.1in
\end{table}
    \section*{Appendix}
    \addcontentsline{toc}{section}{Appendix}  % 如果需要出现在目录中

	%You can have as much text here as you want. The main body must be at most $8$ pages long.
	%For the final version, one more page can be added.
	%If you want, you can use an appendix like this one.  
	
	%The $\mathtt{\backslash onecolumn}$ command above can be kept in place if you prefer a one-column appendix, or can be removed if you prefer a two-column appendix.  Apart from this possible change, the style (font size, spacing, margins, page numbering, etc.) should be kept the same as the main body.

    %\section{A Formal Grammar of Knowledge Equation.}

    %You can have as much text here as you want. The main body must be at most $8$ pages long.
	%For the final version, one more page can be added.
	%If you want, you can use an appendix like this one.  
    
	\section{Ontology Examples.}
    \label{ontology}
The complete ontology comprises only \textbf{40} concepts, \textbf{180} operators, and \textbf{6} thematic topics. In table \ref{tab:ontology}, we present representative samples of core concepts and corresponding operators across topics. The full content can be accessed in our materials.

Our method incorporates the core concepts and operations from these fields. Although some concepts are not directly covered in the ontology library, they can essentially be derived through combinations of our existing operators. This demonstrates the simplicity and expressive power of the knowledge equation.

In the future, we will further expand the coverage of our ontology library to support more complex mathematical reasoning. Our plan primarily includes: \textbf{(1) Completing the ontology library to support geometric and statistical visual reasoning}, and \textbf{(2) Incorporating foundational undergraduate mathematics, such as abstract algebra, mathematical analysis, and topology}. We hope that KE can provide resources and novel insights for future research in autoformalization and theorem proving.

	%The $\mathtt{\backslash onecolumn}$ command above can be kept in place if you prefer a one-column appendix, or can be removed if you prefer a two-column appendix.  Apart from this possible change, the style (font size, spacing, margins, page numbering, etc.) should be kept the same as the main body.

\begin{table}[H]
    \centering
    \caption{Hyperparameters in all the experiments.}
    \label{tab:table 4}
    \vskip 0.15in
    \scalebox{0.9}{\begin{tabular}{llr}
    \toprule
      Type   & Parameter & Value \\ \midrule
      SFT Training & Batch Size & 512 \\
      & Learning Rate & $2.0e-5$ \\
      & Learning Rate Scheduler & Cosine \\
      & Warm-up Ratio & 0.01 \\
      & Optimizer & AdamW \\
      & Epoch & 15\\ \midrule
      Evaluation & Top-p & 0.9 \\
    & Temperature & 0.7 \\
      & Max tokens & 512 \\ \midrule
    \bottomrule
    \end{tabular}}
\vskip -0.1in
\end{table}

\section{Experiments.}

\subsection{Training Details}
\label{appendixB}

We take a fully fine-tuning setting for training DeepSeek-Math-7B-Base as the base model. All training experiments are conducted on 4 NVIDIA A100 GPUs with LLaMA-Factory framework. Detailed hyperparameters utilized for training and evaluation experiments are documented in Table \ref{tab:table 4}.

\subsection{Prompt Details}
\label{prompt}

In this section, we present all the prompts used in this work to facilitate progress. The correspondence between all prompt examples and the tables is shown below.

    \begin{itemize}
        \item Fig \ref{fig:PromptIsabelle}, Fig \ref{fig:PromptI}, Fig \ref{fig:PromptII} and Fig \ref{fig:fewshotpart1}, Fig \ref{fig:fewshotpart2}, Fig \ref{fig:fewshotpart3} present the prompts used to translate natural language to Lean, Coq, and Isabelle respectively, with examples randomly sampled from our parallel dataset.
        \item Fig \ref{fig:prompt4} outlines the prompt used to verify the semantic accuracy of KE during our validation phase.
        \item Fig \ref{fig:prompt5} outlines the prompt used to guide the large language model to perform data synthesis, with templates randomly selected from our template library.
    \end{itemize}
	
	%You can have as much text here as you want. The main body must be at most $8$ pages long.
	%For the final version, one more page can be added.
	%If you want, you can use an appendix like this one.  
	
	%The $\mathtt{\backslash onecolumn}$ command above can be kept in place if you prefer a one-column appendix, or can be removed if you prefer a two-column appendix.  Apart from this possible change, the style (font size, spacing, margins, page numbering, etc.) should be kept the same as the main body.

\begin{figure}
\begin{mdframed}[roundcorner=10pt]
\textbf{Instruction: }  
\\
You are an expert in the Isabelle theorem prover. Your task is to translate theorems from natural language into formal Isabelle statements. Please follow these guidelines:

\begin{enumerate}
    \item Carefully analyze the given theorem in natural language.
    \item Translate it into a correct and precise Isabelle formal statement.
    \item Use the following format for your response: \\
    \quad \texttt{theorem tm\_name :} \\
    \quad \texttt{\ \ fixes \textlangle variable\textrangle} \\
    \quad \texttt{\ \ assumes "\textlangle hypothesis\textrangle"} \\
    \quad \texttt{\ \ shows "\textlangle statement\textrangle"} \\
    \quad \texttt{\ \ sorry}
    \item Focus solely on the translation. Do \emph{not} attempt to prove the theorem or provide additional explanations.
    \item Ensure that your translation accurately captures all the mathematical concepts and relationships expressed in the natural language version.
    \item Use appropriate Isabelle syntax, including correct use of quantifiers, implications, and mathematical symbols.
    \item If the theorem involves specific mathematical structures (e.g., groups, rings, topological spaces), use the corresponding Isabelle definitions and notations.
    \item Do \emph{not} include any proofs, use \texttt{sorry} as a placeholder. Do \emph{not} add any explanations.
\end{enumerate}

The goal is to produce a syntactically correct and semantically accurate formalization in Isabelle. Your translation should faithfully reflect the meaning of the original theorem while following Isabelle conventions and best practices.
\end{mdframed}
\caption{\mbox{Instructions for Translating Natural Language into Isabelle}} 
\label{fig:PromptIsabelle} 
\end{figure}

\begin{figure}
\begin{mdframed}[roundcorner=10pt]
\textbf{Instruction: }  
\\
You are an expert in the Lean4 theorem prover. Your task is to translate theorems from natural language into formal Lean4 statements. Please follow these guidelines:

\begin{enumerate}
    \item Carefully analyze the given theorem in natural language.
    \item Translate it into a correct and precise Lean4 formal statement.
    \item Use the following format for your response: \\
    \quad \texttt{theorem tm\_name : <Lean4 formal statement> := by sorry}
    \item Focus solely on the translation. Do \emph{not} attempt to prove the theorem or provide any explanations.
    \item Ensure that your translation accurately captures all the mathematical concepts and relationships expressed in the natural language version.
    \item Use appropriate Lean4 syntax, including correct use of quantifiers, implications, and mathematical symbols.
    \item If the theorem involves specific mathematical structures (e.g., groups, rings, topological spaces), use the corresponding Lean4 definitions and notations.
    \item Do \emph{not} include any proofs, use \texttt{sorry} as a placeholder. Do \emph{not} add any explanations.
\end{enumerate}

The goal is to produce a syntactically correct and semantically accurate formalization in Lean4. Your translation should faithfully reflect the meaning of the original theorem while following Lean4 conventions and best practices.
\end{mdframed}
\caption{\mbox{Instructions for Translating Natural Language into Lean} } 
\label{fig:PromptI} 
\end{figure}

\newpage

\begin{figure}[H]
\begin{mdframed}[roundcorner=10pt]
\textbf{Instruction: }  
\\
You are an expert in the Coq theorem prover. Your task is to translate theorems from natural language into formal Coq statements. Please follow these guidelines:

\begin{enumerate}
    \item Carefully analyze the given theorem in natural language.
    \item Translate it into a correct and precise Coq formal statement.
    \item Use the following format for your response: \\
    \quad \texttt{Theorem tm\_name : <Coq formal statement>. Proof. Admitted.}
    \item Focus solely on the translation. Do \emph{not} attempt to prove the theorem or provide additional explanations.
    \item Ensure that your translation accurately captures all the mathematical concepts and relationships expressed in the natural language version.
    \item Use appropriate Coq syntax, including correct use of quantifiers, implications, and mathematical symbols.
    \item If the theorem involves specific mathematical structures (e.g., groups, rings, topological spaces), use the corresponding Coq definitions and notations.
    \item Do \emph{not} include any proofs, use \texttt{Admitted} as a placeholder. Do \emph{not} add any explanations.
\end{enumerate}

The goal is to produce a syntactically correct and semantically accurate formalization in Coq. Your translation should faithfully reflect the meaning of the original theorem while following Coq conventions and best practices.
\end{mdframed}
\caption{\mbox{Instructions for Translating Natural Language into Coq}} 
\label{fig:PromptII} 
\end{figure}

\begin{figure}[H]
\begin{mdframed}
\begin{lstlisting}
id: 1 

-- Problem: Find all solutions to the equation \sqrt[3]{3 - \frac{x}{3}} = -2.  

--lean_theorem: 
theorem Unexplored_1 : 
    { (x : ℝ ) | ( 3 - x / 3 ) ^ ( 1 / 3 ) = -2 } = sorry 
    := by sorry 

--coq_theorem: 
Theorem Test_1 :  { x : R} | 1 / (x - 2) < 3 / x } = sorry. 
    Proof. 
    Admitted. 

--Isabelle_theorem: 
theorem Test_1 :  
    shows "{ x :: real . 1 / (x - 2) < 3 / x } = sorry" sorry

id: 2

-- Problem: Given the function f(x) = |1 - 2x| - |1 + x|. Solve the inequality f(x) ≥ 4.

--lean_theorem:
theorem Unexplored_2 : 
    { x : ℝ | |1 - 2 * x| - |1 + x| ≥ 4 } = sorry 
    := by sorry

--coq_theorem:
Theorem Test_2 : 
    { x : R | (Rabs (1 - 2 * x) - Rabs (1 + x)) >= 4 } = sorry.
    Proof.
    Admitted.

--Isabelle_theorem:
theorem Test_2 :
    shows "{x :: real. abs (1 - 2 * x) - abs (1 + x) ≥ 4} = {}"
    sorry

id: 3

-- Problem: If sequence A is an arithmetic sequence with A(1)=3, A(2)=6; find A(5)

--lean_theorem:
theorem Unexplored_3 (A : ℕ → ℝ) 
    (h1 : ∃ d : ℝ, ∀ n : ℕ, A (n + 1) = A n + d) 
    (h2 : A 1 = 3) 
    (h3 : A 2 = 6) : 
    A 5 = 15 := by sorry

--coq_theorem:
Theorem Test_3 (A : nat → R)
    (h1 : exists d, forall n, A (S n) = A n + d)
    (h2 : A 1%nat = 3)
    (h3 : A 2%nat = 6) :
    A 5%nat = 15.
Proof.
Admitted.

\end{lstlisting}
\end{mdframed}
\caption{Few Shots for Translating Natural Language into Formal Language (Part I). } \label{fig:fewshotpart1} 
\end{figure}

\begin{figure}[H]
\begin{mdframed}
\begin{lstlisting}
--Isabelle_theorem:
theorem Test_3 : 
    fixes A :: "nat ⇒ real"
    assumes h1: "∃d. ∀n. A (n + 1) = A n + d"
        and h2: "A 1 = 3"
        and h3: "A 2 = 6"
    shows "A 5 = 15"
    sorry

id: 4

-- Problem: If Set M = {1, 3, 5}, Set N = {2, 3, 4}. Find the union of M and N.

--lean_theorem:
theorem Unexplored_4 (M N : Set ℝ)
    (h1 : M = {1, 3, 5})
    (h2 : N = {2, 3, 4}) :
    M ∪ N = {1, 2, 3, 4, 5} := by sorry

--coq_theorem:
Theorem Test_4 (M N : Ensemble R)
    (h1 : M = [1; 3; 5])
    (h2 : N = [2; 3; 4]) :
    Union M N = [1; 2; 3; 4; 5].
Proof.
Admitted.

--Isabelle_theorem:
theorem Test_4 :
    fixes M N :: "real set"
    assumes h1: "M = {1, 3, 5}"
        and h2: "N = {2, 3, 4}"
    shows "M ∪ N = {1, 2, 3, 4, 5}"
    sorry

id: 5

-- Problem: Solve the following equation: 5(1 - cos x) = 4 sin x

--lean_theorem:
theorem Unexplored_5 : 
    {x : ℝ | 5 * (1 - Real.cos x) = 4 * Real.sin x} = sorry := by sorry

--coq_theorem:
Theorem Test_5 :
    {x : R | 5 * (1 - cos x) = 4 * sin x} = sorry.
Proof.
Admitted.

--Isabelle_theorem:
theorem Test_5 :
    shows "{x :: real. 5 * (1 - cos x) = 4 * sin x} = {}"
    sorry
    
\end{lstlisting}
\end{mdframed}
\caption{Few Shots for Translating Natural Language into Formal Language (Part II). } \label{fig:fewshotpart2} 
\end{figure}

\begin{figure}[H]
\begin{mdframed}
\begin{lstlisting}
id: 6

-- Problem: Given that x, y, z are positive real numbers with product xyz = 1,
-- show that the inequality holds

--lean_theorem:
theorem Unexplored_6 (x y z : ℝ)
    (hx : x > 0) (hy : y > 0) (hz : z > 0)
    (h1 : x * y * z = 1) :
    x^3 / ((1 + y) * (1 + z)) + 
    y^3 / ((1 + z) * (1 + x)) + 
    z^3 / ((1 + x) * (1 + y)) ≥ 3/4 := by sorry

--coq_theorem:
Theorem Test_6 (x y z : R)
    (hx : x > 0) (hy : y > 0) (hz : z > 0)
    (h1 : x * y * z = 1) :
    (x^3 / ((1 + y) * (1 + z)) +
     y^3 / ((1 + z) * (1 + x)) +
     z^3 / ((1 + x) * (1 + y))) >= 3/4.
Proof.
Admitted.

--Isabelle_theorem:
theorem Test_6 :
    fixes x y z :: real
    assumes hx: "x > 0" and hy: "y > 0" and hz: "z > 0"
        and h1: "x * y * z = 1"
    shows "x^3 / ((1 + y) * (1 + z)) + 
           y^3 / ((1 + z) * (1 + x)) + 
           z^3 / ((1 + x) * (1 + y)) ≥ 3/4"
    sorry
    
\end{lstlisting}
\end{mdframed}
\caption{Few Shots for Translating Natural Language into Formal Language (Part III). } \label{fig:fewshotpart3} 
\end{figure}

\begin{figure}[H]
\begin{mdframed}[roundcorner=10pt]
\textbf{Instruction:}
\\
You are an expert in Lean4 language and natural language. When given a math problem described in natural language and a math problem described in Lean4 language, your task is to evaluate the consistency of the two math problems and score them. 

\textbf{Scoring Rules:}
\begin{enumerate}
    \item The full score is 5 points and the lowest score is 0.
    \item When the semantics of all statements of the two math problems are consistent, give full marks of 5 points.
    \item For each inconsistent statement, deduct 1 point until 0 points.
\end{enumerate}

\textbf{Response Format:}
\begin{itemize}
    \item Reply with \texttt{||your points||} in the final sentence
    \item Use the exact "||" format for the score
\end{itemize}

\textbf{Input Format:}
\begin{verbatim}
math problem described in natural language:
<ORIGINAL MATH PROBLEM>

math problem described in Lean4 language:
<LEAN4 MATH PROBLEM>
\end{verbatim}

\textbf{Output Format:}
\begin{verbatim}
<SEMANTIC CONSISTENCY SCORE>
\end{verbatim}
\end{mdframed}
\caption{\mbox{Prompt for Semantic Consistency Judgment}}
\label{fig:prompt4}
\end{figure}

\begin{figure}[H]
\begin{mdframed}[roundcorner=10pt]
\textbf{Instruction:}
\\
You are an expert at creating integrated math problems combining multiple concepts. When provided with knowledge \textbf{K} of operators and concepts, and labeled examples \textbf{E}, your task is to return complex math problems and their labeled results.

\textbf{Rules:}
\begin{enumerate}
    \item \emph{Never use} any new concepts or operators except those in the context!
    \item Do not include any explanatory text.
    \item \emph{Strictly} follow the style of the context.
    \item Combine the provided fragments effectively to create \emph{complex} mathematical problems or proofs.
    \item Return exactly 10 results in the specified format.
\end{enumerate}

\textbf{Output Format:}
\begin{verbatim}
Problem: <problem statement>
Declaration: <required declarations>
Facts: <supporting facts>
Query: <specific question>
\end{verbatim}

\textbf{Input:}
\begin{verbatim}
knowledge K:
## Concepts ##
<EXPLANATION OF CONCEPTS IN K>

## Operators ##
<EXPLANATION OF OPERATORS IN K>

labeled examples E:
<EXAMPLES OF LABELING MATH PROBLEMS>
\end{verbatim}

\textbf{Output:}
\begin{verbatim}
<10 LABELED MATH PROBLEMS>
\end{verbatim}
\end{mdframed}
\caption{\mbox{Prompt for Data Synthesis about Sequence Questions}}
\label{fig:prompt5}
\end{figure}

\newpage

\begin{figure}[H]
\begin{mdframed}
\textbf{MiniF2F T29}. Show that there exist real numbers $a$ and $b$ such that $a$ is irrational, $b$ is irrational, and $a^b$ is rational.
\end{mdframed}
\begin{mdframed}
\begin{lstlisting}
theorem Unexplored_29
    (a : ℝ)(h_a : Irrational a)
    (b : ℝ)(h_b : Irrational b)
    : a ^ b ∈ ℚ 
    := sorry
\end{lstlisting}
\end{mdframed}
\caption{A formalization of MiniF2F T29 in Lean 4. $a ^ b \in \mathbb{Q}$ follows natural language conventions, it constitutes invalid syntax in Lean4. We note that the problem's formalization also contains inaccuracies, though our present focus remains on syntactic errors in this subsection.} \label{fig:MiniF2FT29} 
\end{figure}

\section{Case Study.}
\label{casestudy}
	
This section analyzes common error cases in KELPS translation results, providing a detailed examination of frequent grammatical and semantic issues. Generally, grammatical rules are relatively limited, resulting in a narrower range of error types. In contrast, ensuring semantic consistency often proves more challenging.

\subsection{Syntax Errors}

\textbf{Grammar Errors}. This might be because the model didn’t see enough formal examples, or perhaps there is a discordance between formal language syntax rules and typical organic natural language patterns. A representative example is illustrated in Figure \ref{fig:MiniF2FT29}, where the expression $q \in \mathbb{Q}$, though common and clear in natural language, is not valid in formal syntax. The model may also adopt natural language shortcuts or formats in formal expressions (Figure \ref{fig:FormalMathT452}). This could trace back to the model's exposure to abundant informal content during pre-training, while having relatively limited contact with formalized materials.

\textbf{Type Errors}. These errors primarily occur because, in natural language, a number/object often belongs to multiple types simultaneously, while the informal statement may be insufficient to determine its unique specific type. Generally, Lean4 supports automatic type inference, except for certain specific cases (see Figure \ref{fig:MiniF2FT111}). In contrast, more rigorous formal systems like Coq typically require explicit type declarations in formal statements. Version differences can also trigger such issues. Therefore, file-level (rather than theorem-level) autoformalization is essential for future research.

\subsection{Semantic Errors}

\textbf{Misunderstanding about Informal Statement}. The process of formalizing natural language math problems presents varying levels of difficulty. Since some problems lack explicit mathematical declarations and assumptions, their formalization requires first parsing the natural language semantics and then abstracting mathematical content through modeling. Unfortunately, current mainstream large language models still exhibit limitations in natural language understanding. The representative example in Figure \ref{fig:MiniF2FT198} shows how models lose accuracy when faced with indirect statements. 

\textbf{Misalignment with Informal Statement}. Other prevalent error patterns comprise omission of critical information (Fig \ref{fig:MiniF2FT204}) and inconsistencies with the informal statement (Fig \ref{fig:MiniF2FT155}). These resemble human students' typos in writing solutions, thus being nearly unavoidable. However, these relatively minor errors are theoretically preventable through self-correction and secondary checking.

\begin{figure}[H]
\begin{mdframed}
\textbf{FormalMath T452}. Find all positive integers $a, b, c$ such that $ab + 1$, $bc + 1$, and $ca + 1$ are all equal to factorials of some positive integers. Prove that the answer is: $\boxed{(k! - 1, 1, 1)}$ (and its permutations), where $k \in \mathbb{N}_{>1}$.
\end{mdframed}
\begin{mdframed}
\begin{lstlisting}
Declarations: "a: PositiveIntegers; b: PositiveIntegers; c: PositiveIntegers"

Facts: "a * b + 1 = Factorial(k); b * c + 1 = Factorial(k); c * a + 1 = Factorial(k)"

Query: "(a, b, c) = (k! - 1, 1, 1)",
\end{lstlisting}
\end{mdframed}
\caption{A formalization of FormalMath T452 in Knowledge Equations. While the notation $k!$ clearly denotes the factorial function in human-readable mathematics, it fails to comply with the syntactic requirements of Knowledge Equations. Therefore, it fails to be translated into Lean4. } \label{fig:FormalMathT452} 
\end{figure}

\begin{figure}[b]
\begin{mdframed}
\textbf{MiniF2F T111}. Solve for $a$: $\frac{8^{-1}}{4^{-1}}-a^{-1}=1$. Show that it is -2.
\end{mdframed}
\begin{mdframed}
\begin{lstlisting}
theorem Unexplored_111 
    (a : ℝ)  (h1 : ( 8 : ℝ ) ^ ( -1 ) / ( 4 : ℝ ) ^ ( -1 ) - (a) ^ (-1: ℝ) = 1) 
    : a = -2 := by sorry
\end{lstlisting}
\end{mdframed}
\caption{A formalization of MiniF2F T111 in Lean4. The expression appears almost correct. However, the exponent $(-1)$ in $(8 : \mathbb{R}) ^ {(-1)}$ is not explicitly declared as a real number, causing Lean4 to default its type to $\mathbb{N}$ during computation and resulting in a type mismatch. } \label{fig:MiniF2FT111} 
\end{figure}

\newpage

\begin{figure}[H]
\begin{mdframed}
\textbf{MiniF2F T198}. What digit must be placed in the blank to make the four-digit integer $20\_7$ a multiple of 11? Show that it is 5.
\end{mdframed}
\begin{mdframed}
\begin{lstlisting}
theorem Unexplored_198 
    (d : ℤ)  (h1 : 2057 % 11 = 0) 
    (h2 : 2067 % 11 = 0) (h3 : 2077 % 11 = 0) 
    (h4 : 2087 % 11 = 0) (h5 : d = 5) 
    : 2097 % 11 = 0 := by sorry
\end{lstlisting}
\end{mdframed}
\caption{A formalization of MiniF2F T198 in Lean4. While the generated formal statement is syntactically correct, the model fails to capture the authentic mathematical intent of the problem, consequently producing erroneous results. } \label{fig:MiniF2FT198} 
\end{figure}

\begin{figure}[b]
\begin{mdframed}
\textbf{MiniF2F T204}. Notice that $35 \cdot 40 = 1400$ Find some integer $n$ with $0 \leq n<1399$ such that $n$ is the multiplicative inverse to 160 modulo 1399. Show that it is 1058.
\end{mdframed}
\begin{mdframed}
\begin{lstlisting}
theorem Unexplored_204 
    (n : ℤ)  (h1 : 0 <= n) 
    (h2 : 35 * 40 = 1400) (h3 : n < 1399) 
    : n = 1058 := by sorry
\end{lstlisting}
\end{mdframed}
\caption{A formalization of MiniF2F T204 in Lean4. Although the existing content is correct, the model missed the condition that $n$ is the multiplicative inverse to 160 modulo 1399. This renders the problem unsolvable under the given constraints. } \label{fig:MiniF2FT204} 
\end{figure}

\newpage

\begin{figure}[H]
\begin{mdframed}
\textbf{MiniF2F T155}. If $a$ and $b$ are real numbers, $a^2b^3=\frac{32}{27}$, and $\frac{a}{b^3}=\frac{27}{4}$, what is $a+b$? Show that it is $\frac{8}{3}$.
\end{mdframed}
\begin{mdframed}
\begin{lstlisting}
theorem Unexplored_155 
    (a : ℝ) (b : ℝ)  
    (h1 : a / ( b ^ 3 ) = 27 / 4) (h2 : a * ( b ^ 3 ) = 32 / 27) 
    : a + b = 8 / 3 := by sorry
\end{lstlisting}
\end{mdframed}
\caption{A formalization of MiniF2F T155 in Lean4. The model incorrectly substituted $a$ for $a^2$, resulting in misalignment with the informal statement. } \label{fig:MiniF2FT155} 
\end{figure}

%\section{KELPS DatasSet.}
%\label{KELPSDataset}
%This section talks about the quantity and quality of the KELPS DataSet.

%	\section{Experimentation Details}
%This sections talks about some details about experiments.

	%%%%%%%%%%%%%%%%%%%%%%%%%%%%%%%%%%%%%%%%%%%%%%%%%%%%%%%%%%%%%%%%%%%%%%%%%%%%%%%
	%%%%%%%%%%%%%%%%%%%%%%%%%%%%%%%%%%%%%%%%%%%%%%%%%%%%%%%%%%%%%%%%%%%%%%%%%%%%%%%

\end{document}